\documentclass[a4paper]{article}

\pdfoutput=1
\usepackage[utf8]{inputenc}

\usepackage[super,sort&compress,comma]{natbib}

\usepackage[english]{babel}
\usepackage[T1]{fontenc}
\usepackage{csquotes}

\usepackage[a4paper,top=3cm,bottom=2cm,left=3cm,right=3cm,marginparwidth=1.75cm]{geometry}

\usepackage{amsmath}
\usepackage{graphicx}
\usepackage{hyperref}
\hypersetup{colorlinks=true, allcolors=blue}
\usepackage{authblk}
\usepackage{multirow}
\usepackage{float}

\usepackage{nomencl}
\usepackage{booktabs}
\usepackage{tabularx}
\usepackage{textcomp} 
\usepackage{amsfonts}
\usepackage{outlines} 
\usepackage{subfig}
\usepackage[version=4]{mhchem}
\usepackage{siunitx}
\usepackage{bm}
\usepackage{romannum}

\usepackage[normalem]{ulem} 
\date{}

\DeclareSIUnit\angstrom{\text {Å}}

\usepackage[final]{changes}

\title{Learning Potential Energy Surfaces of Hydrogen Atom Transfer Reactions in Peptides}

\author[1,2]{Marlen Neubert}
\author[1,2]{Patrick Reiser}
\author[3,4,5,*]{Frauke Gräter}
\author[1,2,*]{Pascal Friederich}

\affil[1]{Institute of Theoretical Informatics, Karlsruhe Institute of Technology, Kaiserstr. 12, 76131 Karlsruhe, Germany}
\affil[2]{Institute of Nanotechnology, Karlsruhe Institute of Technology, Kaiserstr. 12, 76131 Karlsruhe, Germany}
\affil[3]{Max Planck Institute for Polymer Research, Mainz 55128, Germany }
\affil[4]{Heidelberg Institute for Theoretical Studies, Heidelberg 69117, Germany}
\affil[5]{Interdisciplinary Center for Scientific Computing, Heidelberg University, Heidelberg 69120, Germany}
\affil[*]{Corresponding author: graeter@mpip-mainz.mpg.de, pascal.friederich@kit.edu}

\begin{document}

\maketitle
\begin{abstract}
    Hydrogen atom transfer (HAT) reactions are essential in many biological processes, such as radical migration in damaged proteins, but their mechanistic pathways remain incompletely understood.
Simulating HAT processes is challenging due to the conflicting requirements of quantum chemical accuracy and biologically relevant time and length scales; thus, neither classical force fields nor DFT-based molecular dynamics simulations are applicable.
Machine-learned potentials offer an alternative, with the ability to learn potential energy surfaces (PESs) that capture reactions and transitions with near-quantum accuracy.
However, training such models to generalize across diverse HAT configurations—especially at radical positions in proteins—requires tailored data generation strategies and careful model selection.
In this work, we systematically generate HAT reaction configurations in peptides to build large datasets using semiempirical methods as well as DFT.
We benchmark three graph neural network architectures, SchNet, Allegro, and MACE, on their ability to learn HAT potential energy surfaces and indirectly predict reaction barriers through direct energy predictions. 
MACE consistently outperforms the other models in energy, force, and reaction barrier prediction accuracy, achieving a mean absolute error of 1.13 kcal/mol on DFT barrier predictions. 
We further perform molecular dynamics simulations and show that the learned potential is stable and reactive under thermal conditions and can serve as an engine for sampling HAT reactions or enhanced-sampling calculations of HAT-free energies.
We show that the trained MACE potential generalizes well beyond our training data by performing out-of-distribution evaluations and analysis of HAT barriers in collagen \Romannum{1} snapshots.
This level of accuracy will enable integration of ML potentials into large-scale collagen simulations to compute reaction rates from predicted barriers, advancing the mechanistic understanding of HAT and radical migration in peptides.
We analyze scaling laws, model transferability, and cost-performance trade-offs, and outline strategies for improvement through the combination of ML potentials with transition state search algorithms and active learning. 
The presented approach is generalizable to other biomolecular systems, offering a method toward quantum-accurate simulations of chemical reactivity in complex biological environments.

\end{abstract}

\section{Introduction}\label{sec:introduction}
Hydrogen atom transfer (HAT) is a central mechanism for radical reactivity and migration in proteins.
We define HAT as the core one-electron step in which a hydrogen atom is abstracted from a donor,
\begin{equation}\label{eq_hat}
     \ce{AH + B^{.} -> A^{.} + BH}
\end{equation}
creating a new radical site on \ce{A} and a closed-shell product \ce{BH}\cite{mayer_2011}.  
Proteins contain radicals both by design in enzymes and by circumstance, due to oxidative or mechanical stress.
In many enzymes, amino acid radicals are deliberately created via one-electron steps using a cofactor or metal center.
The protein guides the created radicals through precise pathways, enabling essential transformations and functions within the enzyme\cite{stubbe_2021}. 
One prominent example of a radical-based enzyme is ribonucleotide reductase, which provides the building blocks of DNA replication and repair.
Outside enzymes, protein radicals are products and propagators of oxidative stress\cite{HAWKINS2001196, davis_2016}.
They are formed by reactive oxygen or nitrogen species and then propagate through the protein via HAT, causing side-chain modifications, cross-linking, and backbone scission. They can also lead to a loss of function and are thus related to several human diseases and the aging process\cite{stubbe_2021}.\\
Load-bearing proteins can form radicals under mechanical stress via homolytic bond scission.
Experiments and simulations on collagen \Romannum{1} tendons under mechanical load reveal the formation of radicals and subsequent production of reactive oxygen species, suggesting that mechanical load contributes to redox signaling and stress\cite{Zapp.2020}. While specific microscopic steps are not fully assigned by Zapp {\it et al.}, HAT, again, provides a plausible route for radical migration among nearby donor/acceptor sites.\\
Whether programmed (enzymes) or stimulus-driven (oxidative, mechanical), protein radicals follow pathways shaped by structure and microenvironment. 
In all cases, HAT reaction barriers and pre-organization of donor/acceptor sites determine kinetics and product distributions.
Due to short lifetimes of HAT, we rely on simulations to understand the fundamentals of radical reactivity and transport in proteins.\\
Recent efforts have focused on approaches that included machine learning (ML).
Riedmiller {\it et al.}\cite{Riedmiller_2024} trained a graph neural network to directly predict HAT barriers in protein environments, focusing on small systems cut out from collagen simulation snapshots. Their model achieved a prediction error of $2.4 \pm 2.5$ kcal/mol on configurations inferred from classical molecular dynamics trajectories and synthetic peptide systems, and an error of $4.6 \pm 4.8$ kcal/mol on out-of-distribution data.
Using the same data, Ulanov {\it et al.}\cite{ulanov_2025} demonstrated that Gaussian process regression models can be employed to predict HAT reaction barriers in low-data regimes, achieving an MAE of $3.23$ kcal/mol.
Both models’ limited prediction accuracy restricts their use in subsequent simulation schemes and the understanding of the HAT reactions themselves. Furthermore, since both approaches directly predict the reaction barriers, the models can not be used in direct MD simulations or enhanced sampling methods.\\
Barrier heights depend on precise knowledge of reaction paths, which in turn require an understanding of the potential energy surface (PES). 
The PES describes the functional relationship between potential energy and atomic positions.
If an accurate PES is known, a molecular system's equilibrium structures or transition states can be found since these correspond to the PES's minima or saddle points. 
When an ML model is directly trained on barrier heights, information on the reaction path and topology of the PES is thus not included.\\
In this work, we model the PESs of HAT reactions in peptides using transferable ML models, which allow us to indirectly predict reaction barrier heights via direct energy predictions. By learning the full PES of HAT reactions in peptide systems, we capture the energetics and forces that govern barrier heights and reaction pathways, and thus can predict more accurate reaction barrier heights.
An accurate model of the PES will also enable further investigations and a deeper understanding of reaction dynamics. The trained ML models representing the PES can also be utilized in MD simulations, enhanced sampling, optimization, and transition state search algorithms, as they are differentiable.\\
Traditionally, there were two approaches to calculating the energy and forces for molecular systems, i.e., the PES.
Ab initio methods, while accurate, are unfeasible for large system sizes due to their computational costs. Classical force fields, instead, are very fast due to their analytical form. The terms in classical force field functions contain many empirical parameters describing bonded and non-bonded interactions (e.g., electrostatic or van der Waals interactions).
However, force fields do not allow bonds to break or form, i.e., no chemical reactions can be simulated. Reactive force fields have been developed to counteract this; however, the accuracy is generally lower than ab initio calculations due to the general empirical approximations \cite{doi:10.1063/1.4909549}.\\ 
Since the PES is a multidimensional function, an analytical expression can also be found by mathematical fitting to data with ab initio accuracy. With the formulation of the construction of the PES as a function approximation problem, it becomes clear where machine learning (ML) methods can be used as efficient tools in this context:
If the relationship between potential energy, including associated analytical derivatives, and atomic positions of a system is constructed using ML methods, the resulting analytical expression is referred to as a machine-learned (ML) potential. 
The training data for ML potentials consists of coordinates as well as elements of all atoms of a system and the corresponding energies and often forces. Since the quality of the resulting ML potential is directly dependent on the quality and quantity of the training data, the latter is typically calculated using an accurate but affordable ab initio method, e.g. density functional theory (DFT).\\
Compared to classical force field methods, ML methods offer the advantage that no constraining assumptions about the functional form of the PES or bonds are needed - the chemical behavior, including long-range interactions and chemical reactions, is learned from the reference data alone \cite{Unke_2021}. ML potentials allow the modeling of the PES of a system with high accuracy and reasonable computational costs. Therefore, their field of application is vast and theoretically ranges over any material in any state, from biomolecules to crystalline systems. Due to its many potential use cases, the development of ML potentials is a very active research field.
Successful ML potentials are based on several different ML architectures, from neural networks to kernel-based methods to graph neural networks (GNNs), with specific advantages and disadvantages \cite{friederich2021machine}.
Regardless of the exact model architecture, training data for ML potentials initially consist of (Cartesian) atom coordinates, the element type of the atoms of a system, and the associated energies in context-dependent configurations. 
Often, information about forces is also part of the training data, as adding them can increase the accuracy of the models and reduce the required training set size \cite{Yao.2018} since there are 3N forces for N atoms instead of just one energy label \cite{Noe.2020}. 
The calculation of energies and forces requires an ab initio method to guarantee the accuracy of the learned potential, but it can also become a bottleneck if a large training data set is needed.\\
The PES exhibits symmetries, which the ML model should also reflect. 
For example, the total energy is invariant if a molecule is translated or rotated, or if two atoms of the same element type exchange.
These invariances can either be explicitly satisfied by choosing a representation of the geometry (e.g. inverse distances), by including them in the functional form of the machine learning model (inductive bias), or by learning them (e.g. through data augmentation).
Currently, the most popular ML model architecture for ML potentials is the graph neural network (GNN), which utilizes the natural graph structure of molecules\cite{10.5555/3305381.3305512,reiser_graph_2022}.  
Since the topology of the molecular structure can be considered as an undirected graph, atoms can be associated with nodes and chemical bonds with edges.
At first, atom feature vectors contain properties such as element types and positions\cite{klicpera2020directional}. 
Information or 'messages' are then exchanged between atoms through message-passing layers, and the model iteratively learns feature representations of the individual atoms' local environments, including information about neighbors and more long-range interactions after several message-passing steps.\\ 
One of the first GNNs to learn PES was SchNet\cite{schutt_schnet_2018}, which is based on invariant convolutions over scalars. 
The model consists of convolutional interaction blocks in which the initial features are updated and the final atom embeddings are learned.
The model ensures rotational invariance of the output by constructing only scalar features and operating on (scalar) interatomic distances, rather than Cartesian atom coordinates.
While SchNet was successfully employed in various chemical applications, the requirement for a lot of ab initio training data was found to be a bottleneck for larger length scales.\\
More recently, equivariant GNNs gained popularity as ML potential models, outperforming previous invariant architectures and displaying higher data efficiency.
Equivariant GNNs can encode more physical information about an atomic system by directly acting on vector quantities while preserving known physical symmetries. 
More specifically, the models are equivariant with respect to transformations under the 3D Euclidean group (rotation, inversion, and translation).
This is relevant for preserving force vectors under rotation of the atomic system.
Equivariance is achieved by not only learning scalar node representations but also higher-order geometric tensor features.\\
Examples of equivariant GNNs include NequIP\cite{batzner_e3-equivariant_2022}, Allegro\cite{musaelian_learning_2023}, and MACE\cite{Batatia2022mace}. 
NequIP utilizes learned scalar and tensor features, and information is propagated via message-passing over relative position vectors. 
While achieving state-of-the-art accuracies on several benchmark datasets, computational performance, specifically training and evaluation speed, constitutes a drawback when scaling to larger systems.
The main disadvantage in this context is the message-passing step since it constructs many neighbors for each atom, hindering the parallelizability of the model.\\
To combat this, Allegro\cite{musaelian_learning_2023}, based on NequIP, learns strictly local equivariant tensor features between edges and employs no message-passing, resulting in an $\mathcal{O}(N)$ scaling with respect to the number of atoms. 
The embedding of the local environment only leads to receptive fields with fixed sizes, which does not reduce accuracy on benchmark datasets.
MACE\cite{Batatia2022mace}, on the other hand, employs a higher-order message-passing strategy to reduce computational costs.
It explicitly models higher-order interactions by constructing many-body features from radial and spherical harmonics basis functions based on the multi-atomic cluster expansion framework\cite{Batatia2022Design}. 
Equivariant messages from these features are then constructed hierarchically via tensor operations.
With this construction method, the authors show that the message-passing can be reduced to two iterations, compared to 4-6 for other equivariant models\cite{batzner_e3-equivariant_2022}.
Evaluations on benchmark datasets show that both Allegro and MACE have high accuracies and good transferability to out-of-distribution data. \\
While benchmark data allows a wide range of comparisons between the latest models, many applications of interest, especially in biochemistry, are often more complex and more specific than common benchmark datasets, and it is not immediately clear which architecture is the most suitable or what kind and how much training data is required.
Thus, trade-offs between training times, data requirements, and accuracy are not intuitively apparent.\\
One of the challenges in the context of HAT in peptides is the increased complexity that a reaction entails.
In addition to training data on equilibrium configurations, global information on the PES, i.e., the reaction's intermediate steps and transition states, is also required.
Accurately training a model thus requires more data, which at the same time needs to be informative to allow a model to capture the increased complexity.
Generating this data constitutes a challenge in itself and requires an efficient data generation workflow. 
This is especially true if not only one specific reaction configuration but, as in our case, various combinations of peptides, radical positions, and reaction paths should be learned.
Higher data requirements to learn the PES of a reaction also mean that we need more ab initio calculations for energy and force labels.
The chosen model must, therefore, be data-efficient. Otherwise, the number of ab initio calculations represents a bottleneck.\\
\begin{figure*}
 \centering
 \includegraphics[width=17.1cm]{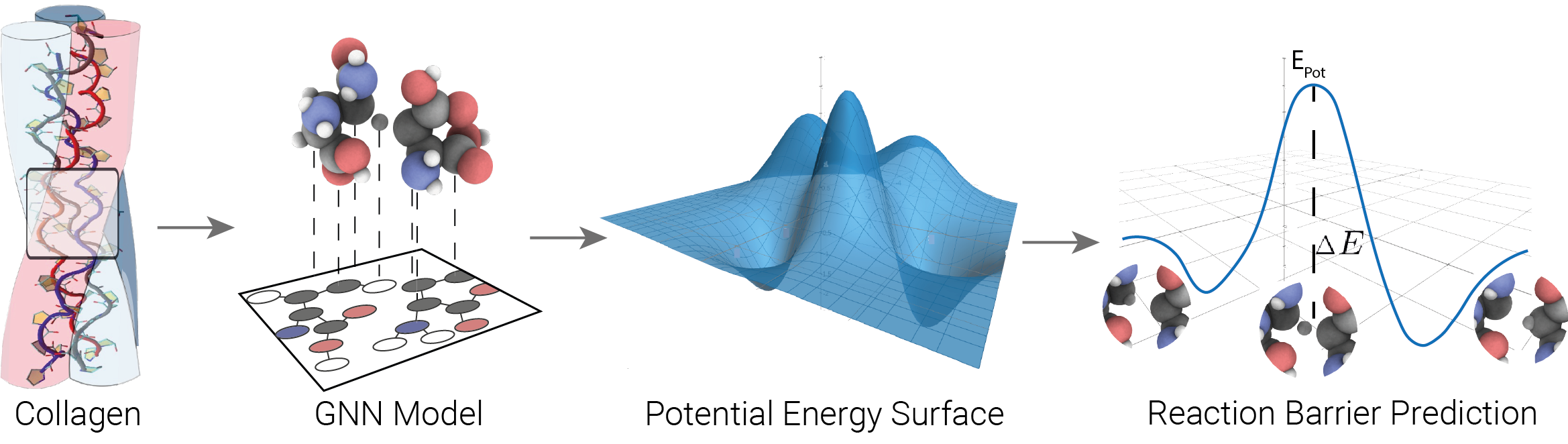}
 \caption{Workflow overview: We generated training data for HAT reactions in peptides and trained graph neural networks to learn the corresponding PES. We used direct energy predictions from these models to indirectly predict HAT reaction barriers.}
 \label{fig:1}
\end{figure*}
In this work, we trained ML potentials to learn potential energy surfaces of HAT reactions in peptides.
Guided by previous studies on collagen and broader protein-radical chemistry, we construct amino-acid and dipeptide systems that incorporate intramolecular (within a peptide) and intermolecular (between two peptides) HAT motifs. 
We developed a workflow to generate training data for reaction configurations in peptides using HAT. Using this workflow, we generated a dataset of 172,000 data points with energy and forces calculated using the semi-empirical method GFN-xTB\cite{bannwarth2020}. 
Additionally, we generated a dataset comprising 125,365 data points at the DFT/bmk/def2-TZVPD level of theory.
We explored the performance of the models SchNet, Allegro, and MACE, estimated a scaling law, and investigated their transferability from small to large systems, both on semi-empirical and DFT data.
We used the trained models to indirectly predict HAT reaction barriers through direct energy predictions (see Figure~\ref{fig:1}). 
The models trained on the PES directly can capture more complexity than previous SOTA direct barrier predictions. Our best MACE model, trained on 65,514 DFT/bmk/def2-TZVPD configurations, achieved an MAE of 1.13 kcal/mol in barrier predictions on out-of-distribution data, compared to a previously reported MAE of $2.4 \pm 2.5$ kcal/mol \cite{Riedmiller_2024}. The significant increase in accuracy achieved here renders the ML model suitable for use in barrier predictions, for example, in radical migration in damaged proteins.
Our molecular dynamics (MD) simulations show that the learned potential is stable and reactive under thermal conditions and can serve as an MD engine for sampling HAT reactions or for enhanced-sampling calculations of HAT-free energies.
We show that our trained ML potential generalizes well beyond our training data by performing out-of-distribution evaluation and analysis of HAT barriers in collagen \Romannum{1} snapshots.

\section{Methods}
\subsection{Data generation} \label{methods_data_generation}
\subsubsection{Training data.} \label{methods_training_data}
\begin{figure}[]
    \centering
    \includegraphics[width=15 cm]{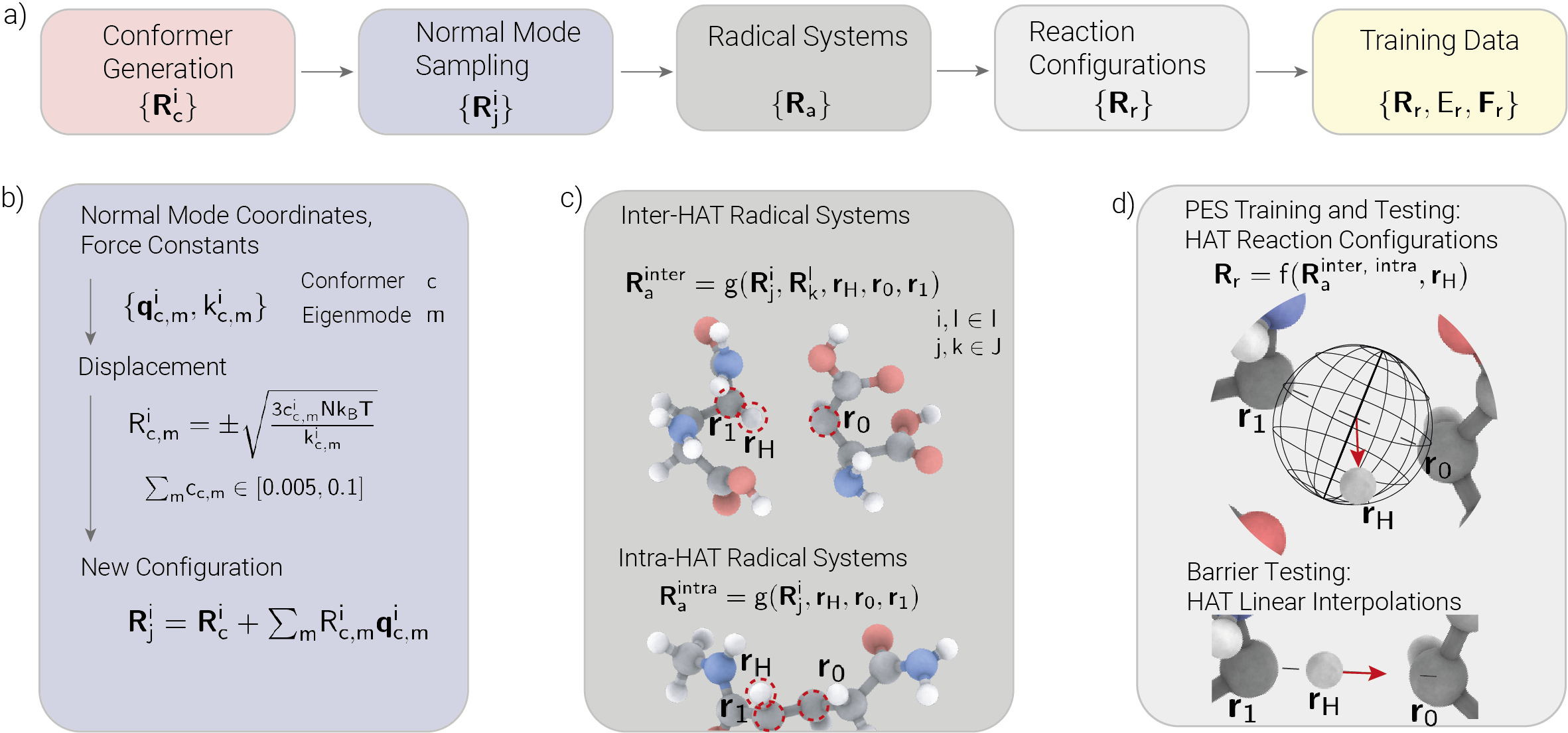}
    \caption{Overview of the training data generation workflow (a). Starting from SMILES representations of amino acids and dipeptides, we generated 3D coordinates using RDKit. After optimizing to get the minimum energy structures, we generated conformers on which we applied normal mode sampling to obtain non-equilibrium structures (b). These configurations serve as input for generating inter- and intra-HAT radical systems (c). Reaction configurations are then sampled by randomly translating the hydrogen atom designated for transfer. Additional evaluation data for the barriers is generated by linear interpolation of the hydrogen atom (d).} 
    \label{fig:2}
\end{figure}

In the following, we present the data generation workflow used to create reaction configurations for HAT in peptides. 
The resulting datasets consist of coordinates and corresponding energies and forces of systems of amino acids and dipeptides.
In addition to equilibrium structures, we must aim to cover a diverse and informative conformational space relevant to describing HAT in various chemical environments.
To achieve this, we combined normal mode sampling and reaction configuration sampling.
The library developed in this work allows us to automatically perform each step of the generation process on many different molecules simultaneously. Each step can also be performed individually, and the framework can, in principle, be adapted to any molecular system.
Figure~\ref{fig:2}a depicts an overview of the steps of the data generation workflow. 
Starting from SMILES, we generated non-equilibrium structures from which we constructed HAT reaction configurations.\\ 
We used RDKit to generate 3D coordinates from SMILES representations\cite{landrum_rdkitrdkit_2024} corresponding to $I$ molecules we want to include in the final training data. We then optimized the initial coordinates using xTB\cite{bannwarth2020} to get minimum energy structures. 
The minimum energy structures were the starting point for a conformer search performed on each with CREST\cite{pracht_automated_2020} (Figure~\ref{fig:2}a).
Since this typically results in numerous structures per molecule, we selected the five lowest-energy conformers and five randomly sampled higher-energy conformers, resulting in $C=10$ conformer configurations $\{\mathbf{R}_c^{i}\}$ per molecule $i$, where $i \in I$ and $c \in C$.\\
We applied normal mode sampling (Figure~\ref{fig:2}b to the chosen conformers in the next step to obtain $J$ physically relevant non-equilibrium structures per molecule $\{\mathbf{R}_j^{i}\}$, where $j \in J$. This allowed us to sample the PES around minima up to a maximum relative energy.
In this step, we distorted the molecules along their normal modes based on methods employed by Rupp {\it et al.}\cite{rupp_machine_2015} and Smith {\it et al.}\cite{Smith.2017, smith_ani-1_2017}.
We used xTB to calculate normal mode coordinates $\mathbf{q}_{c,m}^{i}$ and force constants $k_{c,m}^{i}$ for $m$ eigenmodes of each conformer configuration $c$ per molecule $i$.
The force constants were then used to calculate displacements $R_{c,m}^{i}$ (Figure~\ref{fig:2}b), with which the sampled configuration was generated according to Equation \ref{eq_nms}:
\begin{equation}\label{eq_nms}
    \mathbf{R}_j^{i} = \mathbf{R}_c^{i} + \sum_m R_{c,m}^{i}\mathbf{q}_{c,m}^{i}.
\end{equation}
The normal mode sampled geometries $\mathbf{R}_j^{i}$ are thus superpositions of perturbed normal mode coordinates $\mathbf{q}_{c,m}^{i}$ that pass relative bond length and total energy checks. This ensures that no bonds are broken and that the new configuration's total energy is within a set range.
Normal mode sampling is only an estimation working within the harmonic approximation; in the context of generating training data for ML potentials, it is still beneficial since it allows fast sampling of structures that cover physically relevant PES areas.
The perturbed molecular coordinates $\mathbf{R}_j^{i}$ served as initial structures to build radical systems in the subsequent steps.\\
To create radical systems, we transformed a molecule into a radical by removing a hydrogen atom, creating a radical at position $\mathbf{r}_{0}$ (Figure~\ref{fig:2}c).
We consider two types of radical systems in which reactions occur - intramolecular and intermolecular HAT.
For intra-HAT in peptides, we assume that a transfer occurs within the same molecule, while for inter-HAT, we assume that a hydrogen atom at position $\mathbf{r}_H$ moves between two distinct peptides towards a radical at position $\mathbf{r}_{0}$, thus creating a radical at position $\mathbf{r}_{1}$. 
We implemented a function $g$ that creates the radical systems by performing a selection and geometry modification in the case of inter-HAT systems.
For both system types, the function analyzes given molecules and randomly chooses a hydrogen atom for transfer and a radical position, i.e., the start and end position of the reaction.
The selection function performs distance checks to prevent clashes and includes conditions under which hydrogen atoms can be transferred, depending on the molecule type and atom environment. 
The function generates inter-HAT systems by translating and rotating one randomly chosen molecule and one radical, while the distance between hydrogen atom at $\mathbf{r}_H$ and radical at $\mathbf{r}_0$ is randomly drawn from a $\chi^{2}$-distribution with a maximum distance of 4 \AA. 
The two configurations were arranged so that no clashes occurred, and no other hydrogen atom was closer to the radical position than the atom designated for transfer.
This step results in the creation of radical system configurations $\{\mathbf{R}_a\}$ for both inter- and intra-HAT. 
In the last step, we created reaction configurations from the generated radical systems. A function $f$ modifies the geometry of a system by moving the designated hydrogen atom between the start and end positions.
This displacement function takes a generated radical system $\mathbf{R}_a^{\text{inter, intra}}$ and translates the hydrogen atom $\mathbf{r}_H$ to a point on a sphere with a randomly sampled radius around the center of the start and endpoints of the reaction (Figure~\ref{fig:2}d).
To avoid outlier geometries, we checked again for clashes and energy outliers, i.e., we defined a maximum difference between the minimum energy and the energy of the generated configuration.
This scheme creates a diverse set of reaction systems $\{\mathbf{R}_r\}$ with corresponding energies and forces $\{E_r, \mathbf{F}_r\}$ that differ in geometry, transfer type (intra or inter), type of peptide, as well as the hydrogen and radical positions.
\subsubsection{Evaluation data.} \label{methods_evaluation_data}
To evaluate trained models, we used data generated by the workflow described in Section~\ref{methods_training_data}. We used the generated configurations to directly evaluate the trained models' ability to predict energy and forces.
Due to the randomness in the combinations and finite molecule types we considered, the generated systems can contain the same amino acids or dipeptides as the training data; however, radical and hydrogen atom positions, as well as distances and spatial arrangements of molecules, vary.\\
Since our goal is to predict reaction barriers indirectly using the direct energy predictions from trained models, we generated additional data from linear interpolation of the hydrogen atom at $\mathbf{r}_H$ designated for transfer (Figure~\ref{fig:2}d) similar to what has been done previously in the literature \cite{Riedmiller_2024}.
We generated configurations by moving the hydrogen atom on a linear path between a system's start and end positions with 10 interpolations per system. 
Thus, the reaction barriers for a system were calculated as the energy difference between the highest energy configuration on the path and the start and end positions, respectively.
We applied the linear interpolation scheme to radical systems using the evaluation data, and for additional analysis, to part of the training data.

\subsection{Graph neural networks} \label{methods_gnns}
For all models, we optimized the recommended hyperparameters for our use case while considering training times and computational resources. 
\subsubsection{SchNet.}
We employed the Keras Graph Convolution Neural Networks (KGCNN)\cite{REISER2021100095} implementation of SchNet with a TensorFlow backend, optimizing energy and force predictions.
We used six convolutional interaction blocks with 128 feature dimensions and set the distance cutoff to 5 \r{A}.  
Radial basis function expansion was applied to pairwise distances using 25 Gaussian functions with a distance cutoff of 5 \r{A} and scaling parameter of 0.4 $1/\text{\r{A}}^2$.
The interaction blocks used shifted softplus as the activation function with a pooling method of scatter-sum to aggregate atomic contributions.
The models were trained using the Adam optimizer with an initial learning rate of $10^{-3}$ and loss weights of 1 for energy and 49 for forces.
We applied a linear warmup exponential learning rate scheduler, exponentially decreasing the learning rate by 0.995 per epoch after one warmup epoch.
We used a batch size of 32 and trained the model for 1000 epochs.
We applied an extensive scaler for scaling per-species energies, performing a linear regression to calculate the mean energy per atom type and standard deviation to remove the atomization energy per atom species. 

\subsubsection{Allegro.}
We used the Allegro PyTorch implementation as provided by Musaelian {\it et al.}\cite{musaelian_learning_2023} and built all models with three layers, a radial cutoff of 5 \r{A}, and a latent space with 64 channels.
The radial basis expansion used eight trainable Bessel functions with a polynomial cutoff $p = 6$ and maximum spherical harmonics order $l_{max} = 2$. 
The two-body latent MLP consists of four hidden layers with dimensions [128, 256, 512, 1024], utilizing SiLU nonlinearities and uniform weight initialization.
For the latent MLP, which processes higher-order features, we used three hidden layers with dimensions [1024, 1024, 1024], employing SiLU activations and uniform weight initialization.
A residual connection was applied in the scalar latent space to facilitate efficient propagation of scalar information across layers. The final per-edge energy MLP had a single hidden layer of dimension [128], no nonlinearity, and uniform weight initialization.
We trained all Allegro models using the Adam optimizer with default parameters $\beta_1=0.9$, $\beta_2 = 0.99$, and $\epsilon= 10^{-8}$ without weight decay, using a batch size of 5 and a joint per-atom MSE loss function with weights 1.0 and 1.0 for both energy and forces.
The initial learning rate of 0.001 was reduced by 0.8 using an on-plateau scheduler based on the validation MAE of the energy with a patience of 50.
Early stopping was employed when either the learning rate reached a value of $10^{-6}$, the validation loss did not improve for 50 epochs, or 1000 training epochs were reached.
Per-species energy scaling was applied to normalize atomic energies during training using a Gaussian process regression to compute the mean energy per atom type and standard deviation.

\subsubsection{MACE.}
We employed the MACE PyTorch implementation as provided by Batatia {\it et al.}\cite{Batatia2022mace}, building two-layer models with $l_{max}=2$ in the spherical harmonic expansion, 128 feature channels, and correlation order N=3, i.e., exchanging four-body messages. 
We generated radial features using 8 Bessel basis functions with a polynomial envelope with cutoff p = 5 and set the size of the MLP processing these features for all models to [64,64,64] using SiLu activation functions. 
The readout function performs a linear transformation in the first layer, while the second layer consists of an MLP with a single layer and 16 dimensions.
Models were trained using the Adam optimizer with the AMSGrad variant, with standard parameters $\beta_1=0.9$, $\beta_2 = 0.99$, and $\epsilon= 10^{-8}$.
The learning rate was initially set to 0.005, and an on-plateau scheduler was used to decrease it by a factor of 0.8, with a patience of 50 epochs, based on the validation loss.
For the validation set and final model evaluations, we used an exponential moving average with a decay factor of 0.99.
We used a weighted loss function as described in Batatia {\it et al.}\cite{Batatia2022mace}. 
Initially, the weights for energy and forces were set to 1 and 10, respectively. After 650 epochs, we initiated the second training stage with a reduced learning rate of $10^{-3}$ and a focus on energy loss, with weights set to 1000 for energy and 100 for forces.
We trained all MACE models for 1000 epochs and set the batch size to 5. 
The per-atom energy and standard deviation were calculated using a least-squares regression, which was used to normalize the data during training.

\section{Results and Discussion}
\subsection{Datasets}
We generated datasets for training and evaluating three graph neural network (GNN) architectures, SchNet, Allegro, and MACE, on their ability to predict potential energy surfaces for hydrogen atom transfer (HAT) reactions in peptides. These datasets include both individual reaction configurations and linearly interpolated hydrogen transfer pathways to enable indirect estimation of reaction barriers. All data were generated synthetically through a workflow (see Section~\ref{methods_data_generation}) and calculated both at semi-empirical (xTB) and DFT levels of theory.
A semi-empirical tight-binding model (xTB) was initially used to generate a large and diverse dataset for model development, hyperparameter optimization, and scaling law analysis. In total, we generated 172,042 reaction configurations, of which 45,724 correspond to linear interpolations between hydrogen donor and acceptor positions (see Section~\ref{methods_evaluation_data}).
The scaling law analysis (Section~\ref{results_analysis_gnns}) enabled us to approximate the training set size required to achieve mean absolute errors (MAEs) below 40~meV (1~kcal/mol) for reaction barrier predictions. Based on this analysis, we selected a subset of the xTB dataset for more accurate density functional theory (DFT) calculations. Energies and forces were recalculated at the bmk/def2-TZVPD level using \textsc{Turbomole}, resulting in a DFT dataset with 125,365 configurations, including the full set of 45,724 linear interpolations.

\subsubsection{xTB datasets.}\label{res_xtb_data}
We used the semi-empirical tight-binding method xTB~\cite{bannwarth2020} with an implicit solvent model ($\epsilon = 10$) to compute energies and forces during dataset generation. An epsilon value of 10.0 was chosen to approximate the dielectric environment of collagen (SI~Figure~1).\\
\noindent \textbf{xTB dataset: reaction configurations.} 
SMILES representations of all 20 amino acids and 400 dipeptides (with and without capping groups) were used as the starting point. The capping groups—\ce{NH-CH_3} at the C-terminus and acetyl at the N-terminus—were chosen to mimic the environment of a type I collagen backbone.
To sample relevant conformational space, we applied normal mode sampling to the five lowest-energy and five randomly selected conformers of each molecule. Configurations were retained only if their energy remained within 5.0~eV of the equilibrium structure.
Radical systems for HAT were created by generating all possible combinations of intra- and intermolecular donor–acceptor pairs, including amino acids and dipeptides with and without capping groups. This resulted in a total of 126,318 unique radical systems.
To create intramolecular HAT systems, we equally sampled normal mode sampling configurations across all molecule types(amino acids, dipeptides capped/uncapped). 
We considered all possible pairs of molecule types for intermolecular HAT systems, i.e., HAT between two amino acids, amino acid and dipeptide, two dipeptides (capped and uncapped), all equally weighted.
From each radical system, we generated one reaction configuration by randomly sampling a hydrogen position $\mathbf{r}_\mathrm{H}$ using the method described in Section~\ref{methods_data_generation}. Our preliminary tests showed that including a larger variety of systems improves model generalization more effectively than including multiple configurations per system. As a result, only one configuration—either a start, end, or intermediate hydrogen position—was retained per system, yielding a dataset of 126,318 single-point reaction configurations. System sizes ranged from 15 atoms (e.g., uncapped single amino acids) to approximately 130 atoms (e.g., capped dipeptide–dipeptide pairs). Energy and distance distributions are provided in the SI (see SI Figure~2).
For training and evaluation, the xTB dataset was split into subsets while preserving the distribution of system types (intra- vs. inter-HAT and molecule combinations). The maximum training set size used was 112,191. Detailed dataset statistics and splits are provided in Table~1 in the SI.\\
\noindent \textbf{xTB dataset: linear interpolations.}
To assess the ML models’ ability to reproduce reaction barriers indirectly, we constructed a separate dataset of linearly interpolated hydrogen positions between donor and acceptor atoms, based on radical systems from the xTB dataset. As described in Section~\ref{methods_evaluation_data}, each interpolation consists of 12 configurations (10 intermediate steps plus start and end points). Sampling equally from all system types, we selected 1,861 radical systems from the training data and 2,164 from the test data, resulting in 21,104 and 24,620 configurations, respectively. These datasets were not used for training but only for the evaluation of indirect barrier predictions. 

\subsubsection{DFT datasets.}\label{res_DFT_set}
\begin{figure}[]
    \centering
    \includegraphics[width=15 cm]{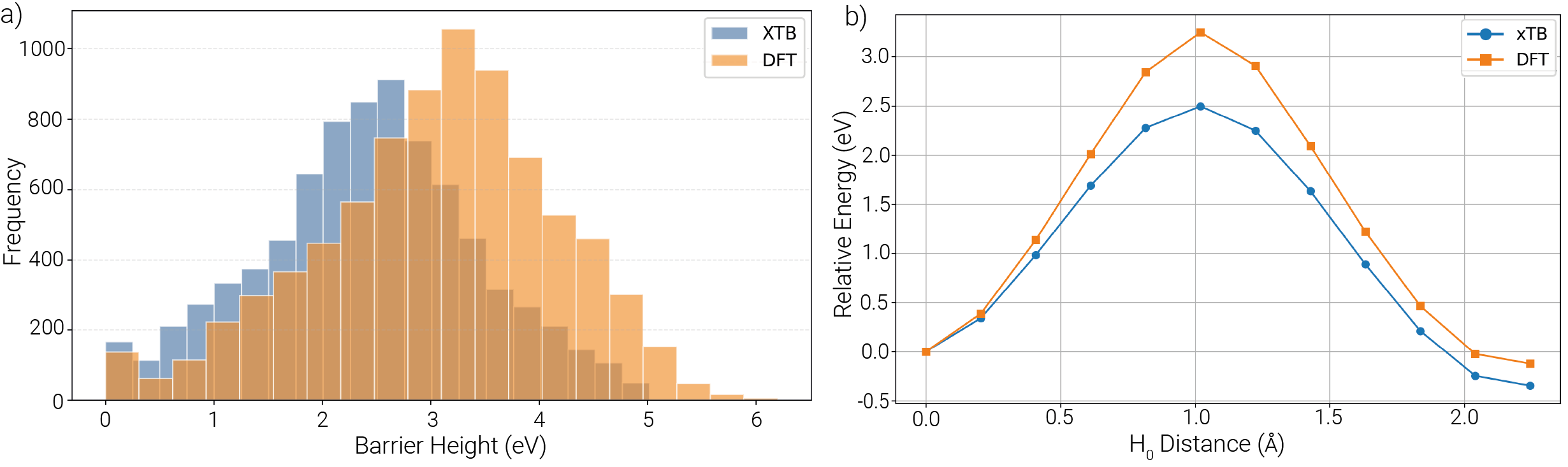}
    \caption{a) Barrier height distributions from linear interpolation test datasets calculated at the xTB and DFT levels. The linear data was only used in test sets, not in training. xTB systematically underestimates barrier heights relative to DFT. b) Example interpolation from the test set for intermolecular HAT between capped Arginine–Glutamate and Lysine–Proline dipeptides (98 atoms). 
    xTB barrier: $\Delta E_\mathrm{left} = 2.50$~eV, $\Delta E_\mathrm{right} = 2.85$~eV; DFT barrier: $\Delta E_\mathrm{left} = 3.25$~eV, $\Delta E_\mathrm{right} = 3.37$~eV.} 
    \label{fig:res_datasets_barriers}
\end{figure}

We recalculated a subset of the xTB data at DFT level using the BMK meta-hybrid functional and the def2-TZVPD basis set, implemented in \textsc{Turbomole}~\cite{ahlrichs_electronic_1989}. BMK was chosen because it was designed for kinetic thermochemistry and has demonstrated good accuracy for reaction barrier heights, including HAT\cite{mangiatordi_2012}, while def2-TZVPD provides diffuse functions beneficial for describing radical and H-transfer transition states.
As with the xTB calculations, we used an implicit solvent model with $\epsilon = 10$ to approximate the aqueous peptide environment. The choice of dielectric constant was informed by testing the sensitivity of barrier heights to $\epsilon$; see SI Figure~1 for details.\\
\noindent \textbf{DFT Dataset: Reaction Configurations.} 
A total of 79,641 single-point configurations were selected from the xTB reaction configuration dataset and recalculated at the DFT level. For consistency, we retained the same configuration indices and data splits across both theory levels. Distribution plots and statistics are provided in the SI Table~1.\\
\noindent \textbf{DFT Dataset: Linear Interpolations.}
The entire xTB interpolation dataset (45,724 configurations) was recalculated at the DFT level, preserving the same system identities and splits (1,861 training, 2,164 test). As shown in Figure~\ref{fig:res_datasets_barriers}, the DFT barriers are consistently higher than those calculated by xTB, indicating that xTB systematically underestimates HAT barrier heights in peptides (see SI Figure~3). 

\subsection{MACE outperforms other graph neural networks}\label{results_analysis_gnns} 
\begin{figure}[]
    \centering
    \includegraphics[width=15 cm]{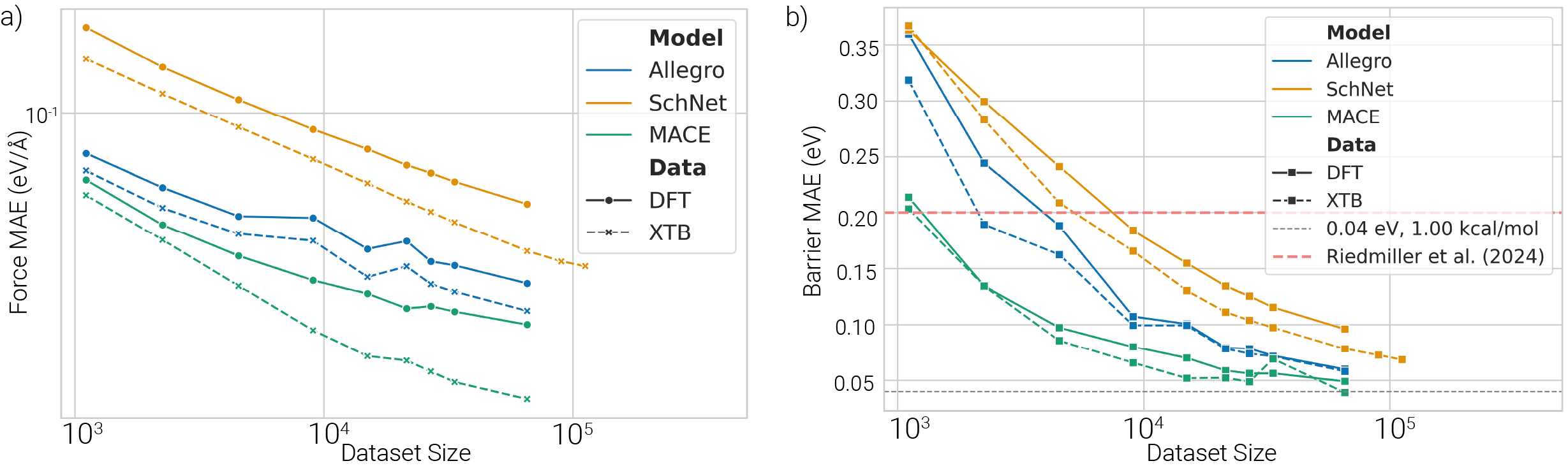}
    \caption{Learning curves of GNNs: a) Test set force MAE vs. training dataset size. b) Test set barrier MAE vs. training dataset size.} 
    \label{fig:res_scaling_law}
\end{figure}

\begin{figure}[]
    \centering
    \includegraphics[width=10 cm]{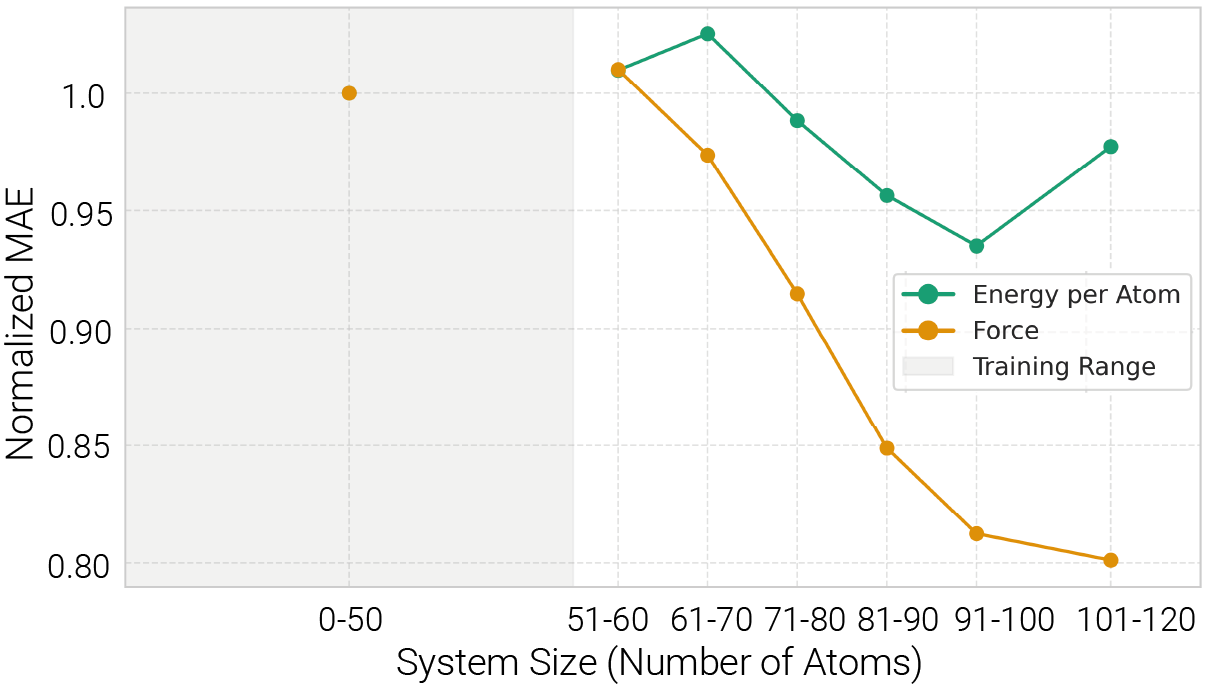}
    \caption{MACE is transferrable to different system sizes: Force MAEs and per-atom energy MAEs vs. atom count. The per-atom energy MAE initially increases, then decreases with increasing system size.}
    \label{fig:res_transf}
\end{figure}

We investigated the three GNN architectures, SchNet, Allegro, and MACE, for predicting energies, forces, and (indirectly) reaction barriers of HAT reactions. 
Our comparison focuses on learning efficiency (scaling laws), transferability to larger systems, and training costs.
assessing their scaling laws, transferability to larger systems, and training efficiency. Note that all models were trained on reaction configurations only, as described in Section~\ref{res_xtb_data} and \ref{res_DFT_set}. Linear interpolation datasets were only used to evaluate the models.
All models were trained on an NVIDIA A100 GPU with 40 GB of memory.\\
\noindent \textbf{Learning curves.}
To assess learning behavior, we trained each model on increasing subsets of both xTB and DFT datasets, using identical configurations, evaluation splits, and model-specific hyperparameters across experiments. 
Learning curves for energy and force MAEs, and therefore also barrier MAEs, decrease as expected with increasing dataset size (Figure~\ref{fig:res_scaling_law}, SI Figure~4a).
Across all dataset sizes, MACE consistently achieves the lowest errors, followed by Allegro, with SchNet showing the highest MAEs. 
When comparing models trained on xTB vs. DFT data, the former consistently exhibits lower errors (Figure~\ref{fig:res_scaling_law}, SI Figure~4b), suggesting that the xTB PESs are inherently easier for the models to learn.
For MACE, the learning curves for xTB and DFT do not run parallel. As the dataset size increases, the gap between the two widens, particularly for force and energy errors. This suggests that DFT-level PESs introduce more complexity, likely requiring higher-order interactions or richer model capacity to be fully captured.\\
\noindent \textbf{Model transferability.} To evaluate model generalization beyond the training distribution, we trained each model on 30,661 DFT configurations with fewer than 50 atoms and tested on a dataset of 34,853 configurations with more than 50 atoms. Performance was also measured on a 3,411-configuration test set of small systems for comparison.
All models exhibit limited transferability to larger systems in terms of energy and barrier MAEs (Table~\ref{tab:model_transf}). Force MAEs, however, remain stable or even improve with increasing system size (Figure~\ref{fig:res_transf}b). MACE again outperforms both Allegro and SchNet across all metrics.
A deeper analysis of MACE on varying system sizes shows that energy MAEs increase with system size, while force MAEs decrease, due to the additive nature of energy prediction errors and stable local force accuracy. For intermediate-sized systems (60–70 atoms), we observe a slight peak in per-atom energy MAEs, followed by a decrease for even larger systems. No clear trend is visible for barrier estimations based on the energy predictions (SI Figure~5).

\begin{table}[ht]
\centering
\caption{Model performance on small ($\leq$ 50 atoms) and large (>50 atoms) DFT test systems. All models were trained on 30,661 small configurations.}
\label{tab:model_transf}
\begin{tabular}{lccc}
\toprule
 & \textbf{Energy MAE} & \textbf{Force MAE} & \textbf{Barrier MAE} \\
\textbf{Model} & \textbf{(meV)} & \textbf{(meV/\AA)} & \textbf{(meV)} \\
\midrule
SchNet ($\leq$50 atoms) & 100 & 74 & 100 \\
Allegro ($\leq$50 atoms) & 60 & 45 & 58  \\
MACE ($\leq$50 atoms) & \textbf{50} & \textbf{33} & \textbf{47}  \\
\midrule
SchNet (>50 atoms) & 234 & 71 & 146 \\
Allegro (>50 atoms) & 120 & 44 & 94  \\
MACE (>50 atoms) & \textbf{100} & \textbf{31} & \textbf{66}  \\
\bottomrule
\end{tabular}
\end{table}

\begin{figure}[]
    \centering
    \includegraphics[width=15 cm]{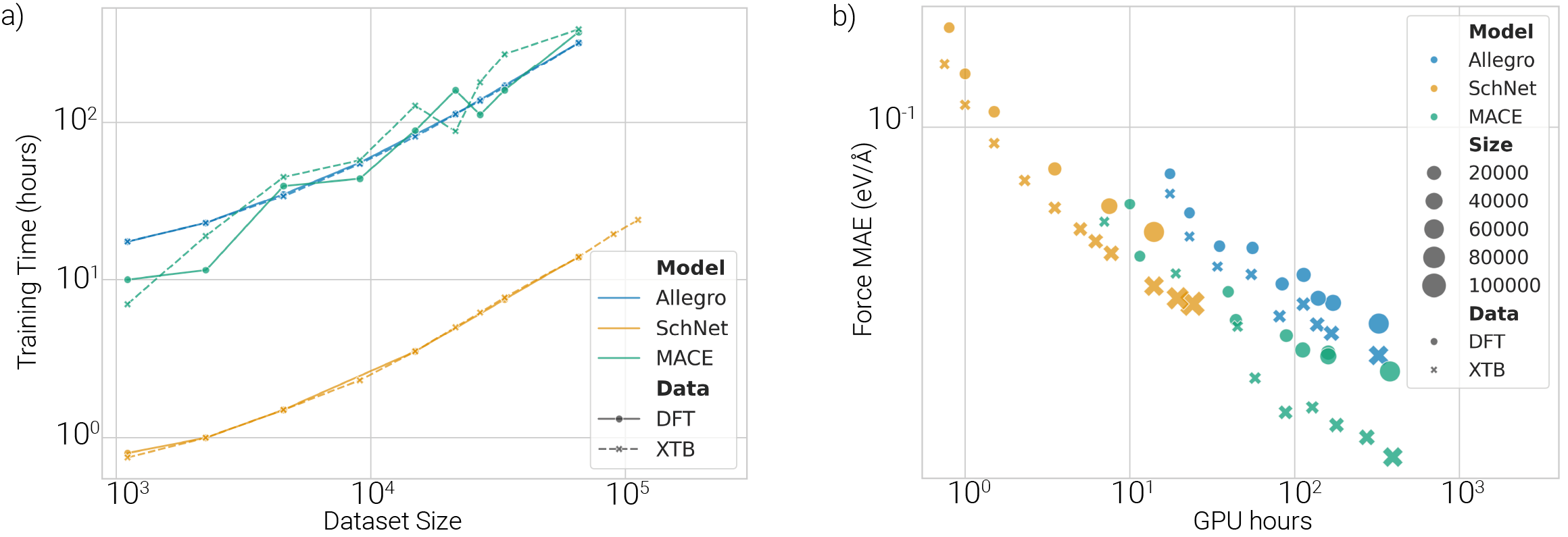}
    \caption{a) Training time vs. dataset size. Training times increase with dataset size. Allegro and MACE need substantially longer training times than SchNet. b) GPU hours vs. test force MAE. MACE and Allegro are more data-efficient but require significantly more compute.} 
    \label{fig:res_training_times}
\end{figure}

\noindent \textbf{Training efficiency and resource trade-offs.}
Training times for Allegro and MACE are substantially longer than for SchNet across all dataset sizes (Figure~\ref{fig:res_training_times}a). MACE requires up to 20 times more GPU hours than SchNet, but achieves comparable or better force accuracy with only half the data (Figure~\ref{fig:res_training_times}b), highlighting a trade-off between data efficiency and computational cost.

\subsection{Final DFT model training}
\begin{figure}[]
    \centering
    \includegraphics[width=15 cm]{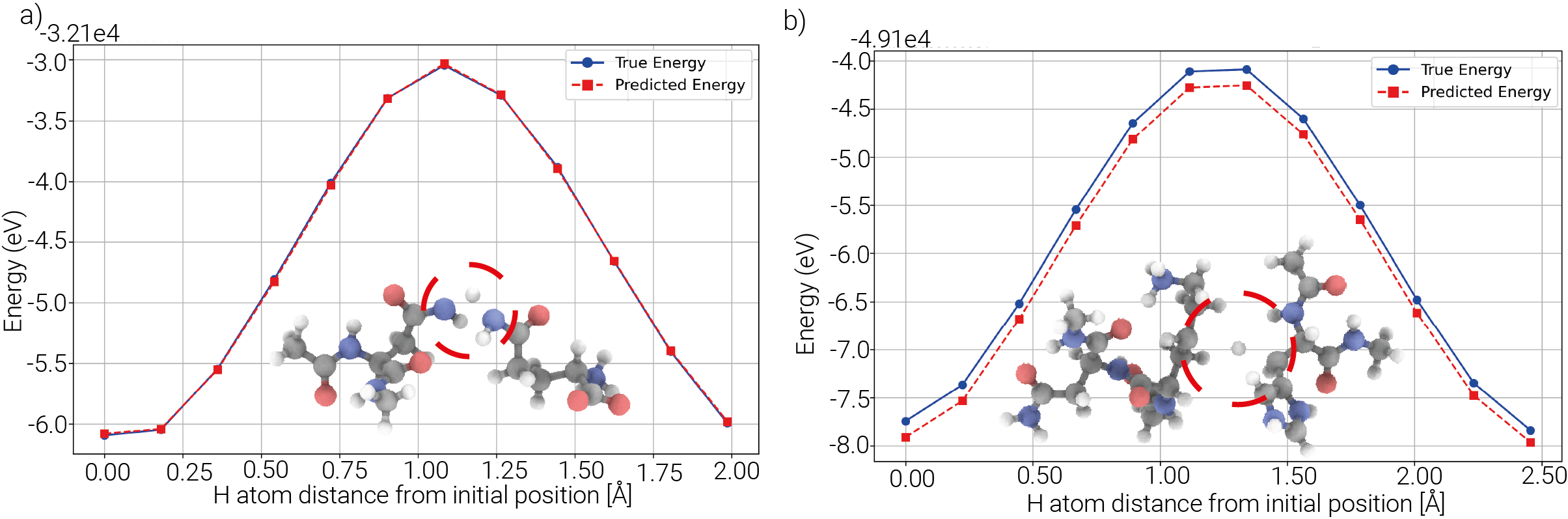}
    \caption{Barrier predictions with MACE:
    a) Inter HAT system comprising aspartate and alanine with capping groups (45 atoms). Very good agreement between predicted and true single point energies results in low barrier errors $|\Delta E_{left}|=1.5 meV, |\Delta E_{right}|=0.82 meV$ b) Selected inter HAT system comprising a lysine-asparagine dipeptide and histidine with capping groups (77 atoms) with high energy and low barrier errors. 
    Error cancellation yields accurate barrier estimates despite offset energy predictions.
    $|\Delta E_{left}|=2.3 meV, |\Delta E_{right}|=41 meV$. } 
    \label{fig:res_barriers}
\end{figure}

To obtain final models trained on all available data, we trained SchNet, Allegro, and MACE on 65,514 DFT configurations and tested on 6,836 unseen configurations. 
Reaction barriers were derived from direct energy predictions for 2,164 HAT test systems, comprising 24,620 single-point evaluations.
MACE achieves the best performance, with the lowest energy (68~meV), force (28~meV/Å), and barrier (49~meV) MAEs (Table~\ref{tab:model_comparison}). Despite higher errors in energy predictions, all models exhibit lower errors in barrier predictions, likely due to systematic error cancellation when models systematically over-/underestimated energies.
In some cases, predicted energy profiles closely match DFT reference values (Figure~\ref{fig:res_barriers}a), while in others, consistent over- or underestimation across the pathway leads to accurate relative energies and barriers (Figure~\ref{fig:res_barriers}b).
\begin{table}[ht]
\centering
\caption{Test error of final models trained on 65,514 DFT configurations and tested on 6,836 unseen configurations and 2,164 barrier evaluations.}
\label{tab:model_comparison}
\begin{tabular}{lccc}
\toprule
&\textbf{Energy MAE}& \textbf{Force MAE} & \textbf{Barrier MAE} \\
\textbf{Model} &(meV)& (meV/\AA) & (meV) \\
\midrule
SchNet & 97 & 58 & 96 \\
Allegro & 79 & 36 & 60  \\
MACE & \textbf{68} & \textbf{28} & \textbf{49}  \\
\bottomrule
\end{tabular}
\end{table}

\subsection{Molecular dynamics simulations}
We assessed whether our learned PES (MACE, trained on DFT data) is stable enough to sustain molecular dynamics simulations and to observe HAT events across a set of 30 unseen test systems.
To do so, we integrated the trained model into ASE as a calculator that provides energies and forces for each step of the NVT simulation.
We considered three settings: unbiased dynamics, medium steering, and strong steering. Steering applies a one-dimensional bias along the donor-acceptor transfer coordinate, using a harmonic restraint that is linearly shifted from an initial to a target value over a set number of steps.
For each system, we launched five independent replicas with randomized velocity seeds and trajectory lengths of 100 ps (unbiased) and 50 ps (steered).
Detailed configurations for the simulations and per-system results are provided in the SI4, table SI3.
We considered a successful HAT transfer if the transferring H-atom's nearest heavy-atom neighbor switches from the original donor to the intended acceptor and remains in this state until the end of the simulation.
Across 30 unseen HAT test systems, HAT occurred spontaneously in 6 systems under unbiased dynamics (success probability by the end $P_\mathrm{end}=0.10$), in 19 systems under medium steering ($P_\mathrm{end}=0.51$), and in 24 systems under strong steering ($P_\mathrm{end}=0.65$). 
All trajectories maintained stable total energies and temperatures throughout the simulations. In two systems, a short initial temperature spike was observed in 1 out of 5 replicas each, which relaxed within < 0.5 ps and did not affect the subsequent dynamics.
These results confirm that the learned MACE potential is stable for finite-temperature dynamics and can both sustain and promote HAT events under moderate biasing.

\subsection{Out of distribution evaluation}
To test whether our DFT-trained ML potential generalizes beyond our generated data, we evaluated it on HAT systems extracted from collagen \Romannum{1} MD snapshots. This out-of-distribution (OOD) set probes new local chemistry, more irregular environments, and realistic finite-temperature geometries.
From two snapshots, we identified HAT donor–acceptor candidates and, for each, constructed matched parent–child pairs that preserve the same H-transfer to investigate the effects of size and context on the barriers. For intra-HAT, the parent is a capped tripeptide, and the child is the corresponding capped dipeptide obtained by removing one residue from the parent. For inter-HAT, the parent is a pair of one capped dipeptide and one capped amino acid, and the child removes one amino acid from the dipeptide, leaving the D–H–A identity unchanged. We added the same capping groups as for our training data (N-terminus acetyl and C-terminus—\ce{NH-CH_3} groups). 
For each system, we generated a 12-image linear H-transfer path (10 intermediates plus endpoints) and defined left/right barriers as the path maximum relative to the start/end states, as in our evaluation protocol. We computed single-point energies and forces at the BMK/def2-TZVPD level to match the trained model's level of theory.
The final OOD benchmark contains 148 HAT systems (74 intra, 74 inter), spanning 28–75 atoms.\\
Barrier analysis across 74 paired full (parent) and reduced (child) peptide systems reveals that the parent and child barrier distributions largely overlap. The median child–parent shift is ~0, and mean per-pair deviations are small (0.08 eV). The same conclusion holds when splitting into intra vs. inter subsets.
Smaller capped systems that keep only the residues directly connected to the donor and acceptor yield barrier estimates comparable to those of the full systems, with only minor case-by-case differences.\\ 
Our trained MACE model achieves a barrier MAE of 76 meV over all systems, an energy MAE of 331 meV, and a force MAE of 47 meV/\AA. 
After excluding 22 outlier systems, the barrier MAE decreases to 72 meV, the energy MAE to 90 meV, and the force MAE to 38 meV/\AA. 
The flagged systems all show a near-constant energy offset along the reaction path, indicating that most of their energy error stems from system-specific constant shifts rather than from errors in the shape of the reaction profile.
Compared to the synthetic data, the errors are higher in absolute terms, as expected from an OOD test, but barriers remain comparatively robust due to cancellation of systematic offsets within each trajectory.
Larger parent and smaller child systems exhibit similar barrier MAEs (76 meV each), indicating that the model retains its barrier accuracy even when the environment is increased, as long as the local H-transfer motif is preserved.
On realistic systems derived from collagen \Romannum{1} snapshots, our trained MACE model thus preserves barrier shapes well and achieves near-in-distribution barrier accuracy. Once outliers with per-system offsets are removed, it also approaches near-in-distribution energy and force errors.

\subsection{Discussion}
We developed a pipeline for generating xTB and DFT-labeled configurations of HAT reactions in peptides and used it to investigate three ML potential architectures: SchNet, Allegro, and MACE.
The xTB data served us in our initial tests to estimate sufficiently large training datasets and allowed us to compare learning difficulties between a semi-empirical PES and a DFT PES. 
Beyond xTB, pretrained universal ML potentials could be used in the exploration stage to propose reaction configurations and approximate labels. 
Our present peptide-radical HAT and implicit-solvent setting demands open-shell coverage and domain alignment that are not yet guaranteed in foundation models, and xTB remains highly competitive at our system sizes. Since our key findings are based on DFT re-labeling, we did not investigate foundation-model baselines in this study.\\
We used an implicit solvent model for all calculations. Explicit water and heterogeneous protein environments are expected to modulate barrier heights. In this work, we tested barrier sensitivity and selected a dielectric constant value to approximate the peptide environment. To mitigate this limitation, future datasets could include explicit solvent shells and counter-ions around the reactive site during DFT labelling. At the model level, ML/MM approaches can be employed to capture environmental responses while maintaining DFT-level accuracy at the HAT center.\\
MACE consistently outperforms the other models regarding energy, force, and barrier MAEs but is also the most computationally demanding. Allegro achieves slightly lower accuracy and has comparable training costs. SchNet trains quickly but suffers from higher prediction errors, especially when the dataset size as well as the budget for training compute is large.
Models trained on xTB data achieve lower errors than their DFT-trained counterparts, suggesting that xTB PESs are inherently easier to learn, at least for the HAT/peptide systems investigated here. 
The approximate nature of the xTB PES might reduce the complexity of the PES compared to DFT. DFT-trained models may better reflect physical reality despite higher energy and force errors.
As dataset sizes increase, this difference becomes more pronounced, particularly for MACE, indicating that DFT PESs likely require more complex models or additional model capacity (e.g. higher-body terms).
Though hyperparameter searches were conducted for both data types, they were limited to small datasets due to training costs. Improved performance on DFT data might be achievable with more extensive hyperparameter tuning on larger datasets.\\
Our MD experiments demonstrate that the learned MACE–DFT potential is dynamically stable and supports reactive HAT at finite temperature. Under unbiased NVT dynamics, spontaneous HAT events occurred in a minority of test systems, whereas physically motivated steering increased the event frequency. 
This suggests that the model can serve as an engine for enhanced-sampling calculations of HAT-free energies. The steering coordinate $q$ used here could be reused directly as the collective variable for umbrella windows or as the bias in well-tempered metadynamics.      
It is expected that some regions of configuration space will be poorly sampled, e.g. near transition states. Incorporating an active-learning strategy that monitors model uncertainty during MD and triggers targeted DFT re-labeling in those regions could improve sampling, reduce bias, and accelerate convergence of the free-energy estimates.\\
Energy prediction transferability to larger systems is challenging for all model architectures we investigated, likely because global, extensive properties such as total energy suffers from additive errors. 
The relatively stable or even improved force MAEs suggest that models generalize well locally, and that force predictions benefit from larger local environments or averaging effects in bigger systems.
However, very large systems may introduce long-range interactions that are not present during training, further complicating energy and force predictions. Since SchNet, Allegro, and MACE rely on local atomic environments, transferability of energies depends on the presence and diversity of long-range effects in the training data.\\
Compared to our in-distribution tests, the collagen \Romannum{1} MD-snapshot evaluation is an explicit out-of-distribution stress test introducing finite-temperature strain, context truncation, and a shifted geometry distribution.
Our generated training data spans all residues in idealized contexts, whereas the collagen snapshots sample a subset of chemistries in specific packed environments and sit off minima with thermally distorted bonds, angles, and torsions.
Despite this mismatch, our model achieves a barrier MAE of 76 meV on the full extracted snapshot set, and overall barrier shapes and TS locations are retained.
Our analysis of reaction barriers in parent/child systems indicates that barriers remain essentially unchanged when trimming the environment. Larger parent and smaller child systems exhibit similar barrier MAEs, highlighting size transferability and showing that we can, in practice, use smaller capped systems as long as the local H-transfer motif and its immediate contacts are preserved.
A small number of matched parent–child systems were flagged as outliers, all of which exhibit a nearly constant offset between predicted and DFT energies along the reaction path. These outliers occur in both intra and inter HAT motifs and span neutral as well as charged residue combinations, indicating that they are not confined to a single topology or simple residue type. Instead, the model reproduces the barrier shape for these cases but assigns an incorrect absolute energy reference to the capped fragment, consistent with a fragment-specific calibration error rather than a failure to describe the reaction coordinate itself. In practice, these cases dominate the global energy MAE but have limited impact on barrier errors, since the constant offset largely cancels along the path.\\
Applying MACE at the collagen scale could be made possible using hybrid schemes rather than single end-to-end models. One option could be ML/MM embeddings that treat only the reactive HAT region at ML accuracy, while the protein/solvent environment remains classical. Alternatively, the ML potential could be used as a barrier emulator within large-scale kinetics (e.g., kinetic Monte Carlo \cite{rennekamp_hybrid_2020}) driven by on-the-fly computed ML barriers along representative pathways.
For small datasets (<10k configurations), SchNet offers a practical trade-off between training time and accuracy. In some cases, if a limited amount of resources is available, it might be more efficient to train more data on SchNet for fewer GPU hours (Figure~\ref{fig:res_training_times}b). For larger datasets (>30k configurations), MACE becomes more advantageous due to its data efficiency, despite longer training times. Allegro falls in between in terms of both cost and performance. 
All models were trained on a single GPU for consistency, but MACE in particular supports parallel training, which could significantly reduce training time. 
All models achieve more accurate barrier predictions than direct energy predictions, likely due to error cancellation within reaction pathways. 
Overall, our results show that machine-learned potentials can accurately predict DFT-level reaction barriers from direct energy predictions, with MACE providing the most reliable and generalizable performance across the tasks considered.
In our tests, we used linear interpolations for barrier estimations, which need to be refined via optimization and transition state searches to get more accurate barriers. 
However, our scheme of indirectly predicting the reaction barriers should work equally well when using refined reaction configurations.
More accurate barriers could also be obtained via transition state searches using the trained models directly, since they provide a cheaper way to get Hessians via auto-differentiation. As MACE provides differentiable energy landscapes, it could also be integrated into such optimization schemes. To do so, we would need further tests to ensure accurate Hessian predictions, but initial investigations already suggest this works well\cite{yuan_analytical_2024}. 
Finally, combining our pretrained models with transition state search algorithms is also very well suited to be used in an active learning approach to retrain models on relevant PES regions for HAT reactions.

\section{Conclusion}
In this work, we developed a workflow for generating datasets and training machine learning potentials for HAT reactions in peptides.
We trained and assessed the performance of three GNN architectures, SchNet, Allegro, and MACE, on both semi-empirical (xTB) and DFT-level potential energy surfaces.\\
MACE consistently outperforms SchNet and Allegro in energy, force, and reaction barrier prediction accuracy, albeit at a higher computational cost. Our best MACE model achieved a mean absolute error of 1.13 kcal/mol in indirectly predicting DFT-calculated HAT reaction barriers, substantially improving upon previous machine learning approaches. 
This accuracy is critical for reliable reaction rate predictions, as errors in barriers propagate exponentially into rate constants. 
Our approach leverages direct energy predictions to model complex PESs and estimate reaction barriers without the need for explicit transition-state data, offering a generalizable alternative to direct barrier prediction schemes.
Our analysis of scaling laws, transferability, and training costs highlights the importance of balancing model complexity, dataset size, and computational resources. While xTB PESs are easier to learn and allow fast prototyping, they may limit generalization to high-accuracy regimes. In contrast, DFT-trained models better reflect physical reality and are essential for robust and transferable predictions.\\
The trained models show good transferability of force predictions across system sizes, though total energy errors grow with system size, likely due to a lack of out-of-distribution generalization to larger systems and missing long-range interactions.\\
Our MD experiments show that the learned potential is stable and reactive under thermal conditions and can serve as an MD engine for sampling HAT reactions.
Umbrella sampling along the H-transfer coordinate or metadynamics on the same CV could be the next steps for obtaining HAT free-energy profiles at MACE cost. Coupled with an active-learning loop that identifies high-uncertainty configurations for selective DFT re-labeling, this workflow could deliver accurate free-energy surfaces.
The trained MACE model is also suitable as an emulator, for example, in kinetic Monte Carlo schemes to invoke reactions within a protein.  
We evaluated the DFT-MACE potential on collagen \Romannum{1} MD–derived HAT snapshots, which is an explicit out-of-distribution setting relative to our benchmark, to test transfer to a realistic peptide environment.
Our MACE potential generalises well beyond its synthetic training set. It preserves barrier shapes and TS locations and achieves barrier errors that remain within a few 10 meV of its in-distribution performance. The similar accuracy on larger parent and reduced child systems demonstrates size transferability, enabling the use of compact capped fragments without sacrificing barrier fidelity. The few remaining outliers are dominated by fragment-specific constant energy offsets rather than distorted reaction profiles, pointing to clear avenues for improvement via targeted retraining or simple fragment-level corrections that reduce system-specific shifts while retaining the already robust barrier predictions.
Transition-state optimization with ML-predicted Hessians could further refine barrier predictions. 
An active learning strategy that combines pre-trained models with automated transition state searches may help systematically improve accuracy and broaden the configurational diversity of training data \cite{zhou2024pal}.\\
The presented workflow is not limited to HAT in peptides and can be translated to other reactive processes in biomolecular systems. As ML potentials continue to mature, they offer a path toward simulating complex chemical reactivity in biologically and chemically realistic environments, bridging the gap between quantum accuracy and large-scale dynamics.

\subsection*{Data availability}
The training data, trained models, as well as the code to train all machine learning models and the scripts to reproduce the results of this paper, can be found on
\url{https://github.com/aimat-lab/hat_pes_learning} (v1.0) and on Zenodo (\url{https://doi.org/10.5281/zenodo.17670180}).


\subsection*{Author contributions}
All authors contributed to the idea and the preparation of the manuscript. 
M.N. implemented the methods and conducted the computational experiments.

\subsection*{Conflicts of interest}
There are no conflicts of interest to declare.

\subsection*{Acknowledgements}
The present contribution is supported by the Helmholtz Association under the joint research school “HIDSS4Health – Helmholtz Information and Data Science School for Health”.\\ 
We acknowledge funding from the Klaus Tschira Stiftung gGmbH (SIMPLAIX Project 1, to P.F. and F.G.) and the European Research Council (ERC) under the European Union’s Horizon 2020 research and innovation programme (grant agreement No. 101002812) (to F.G.).
This work was performed on the HoreKa supercomputer funded by the Ministry of Science, Research and the Arts Baden-Württemberg and by the Federal Ministry of Education and Research.
The authors acknowledge support by the state of Baden-Württemberg through bwHPC.

\printnomenclature

\clearpage



\bibliography{bib}
\bibliographystyle{rsc} 

\newpage
\begin{center}
    \LARGE Supporting Information
\end{center}
\vspace{1cm}
\appendix
\setcounter{section}{0}
\renewcommand{\thesection}{SI \arabic{section}}
\setcounter{figure}{0}
\renewcommand{\thefigure}{SI \arabic{figure}}
\setcounter{table}{0}
\renewcommand{\thetable}{SI \arabic{table}}

\section{Datasets}

\begin{table}[H]
\centering
\caption{Summary of dataset composition used, listing the number of molecular configurations available at the xTB and DFT levels of theory. The datasets are divided into training, evaluation, and test sets for both single molecular systems and linear interpolation tasks.}
\label{tab:model_comparison_SI}
\begin{tabular}{lcc}
\toprule
\textbf{Dataset type} & \textbf{xTB} & \textbf{DFT} \\
\midrule
Total & 172,042 & 125,365 \\
Single Systems Training & 112,191 & 65,514  \\
Single Systems Evaluation & 7,291 & 7,291  \\
Single Systems Test & 6,836 & 6,836 \\
\addlinespace
Linear Interpolation Evaluation & 24,620 & 24,620  \\
Linear Interpolation Test & 21,104 & 21,104  \\
\bottomrule
\end{tabular}
\end{table}

\begin{figure}[H]
    \centering
    \includegraphics[width=15 cm]{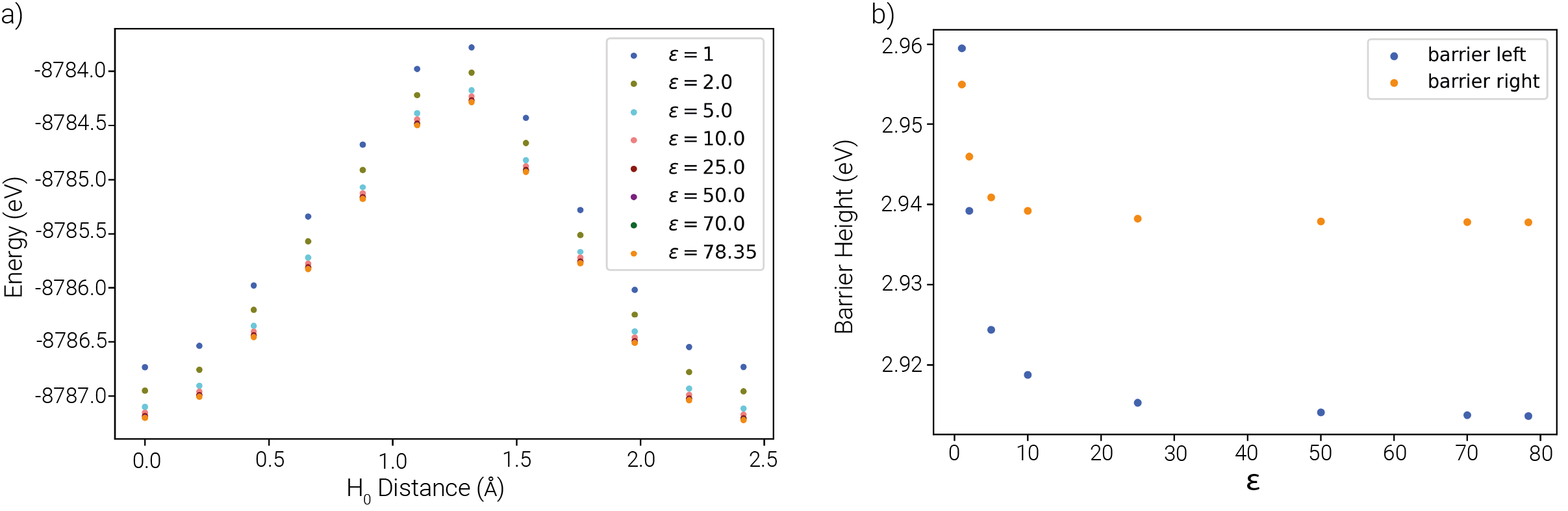}
    \caption{ Dielectric constant tests for implicit solvent calculations. The dielectric constant ($\epsilon$) used in xTB and DFT calculations was selected based on convergence behaviour of reaction barrier heights and considerations of typical protein environments. a) Example energy profile showing that the barrier height stabilized for $\epsilon >10.0$. b) Barrier height as a function of $\epsilon$, illustrating convergence beyond $\epsilon =10.0$. } 
    \label{fig:SI_epsilon_tests}
\end{figure}

\begin{figure}[H]
    \centering
    \includegraphics[width=15 cm]{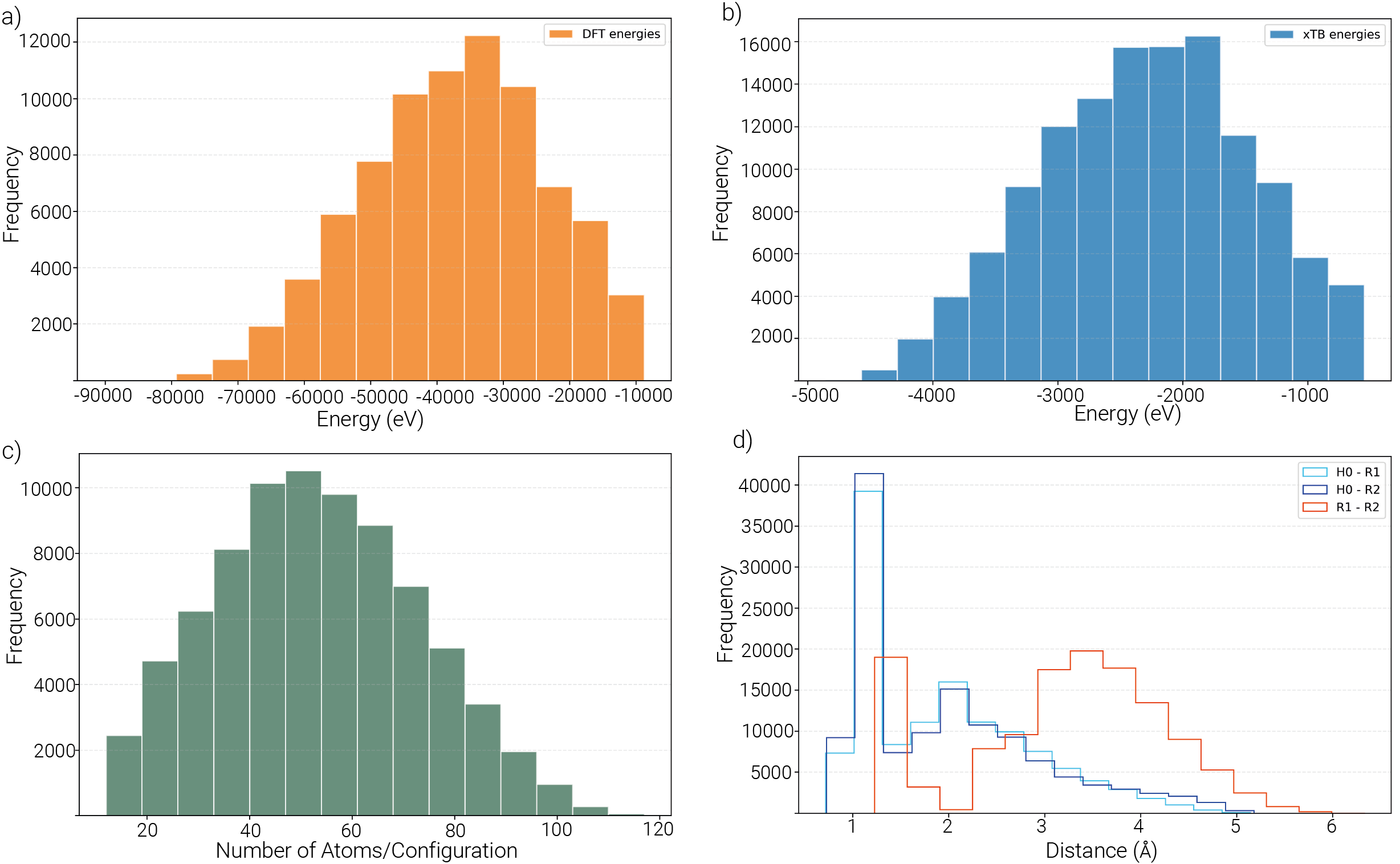}
    \caption{DFT and xTB dataset statistics. Potential energy distributions of the configurations of all a) DFT and b) xTB data. Distribution of c) the number of atoms per configuration and d) the Hydrogen atom transferred - radical distances within the DFT dataset. } 
    \label{fig:SI_dataset_statistics}
\end{figure}

\begin{figure}[H]
    \centering
    \includegraphics[width=15 cm]{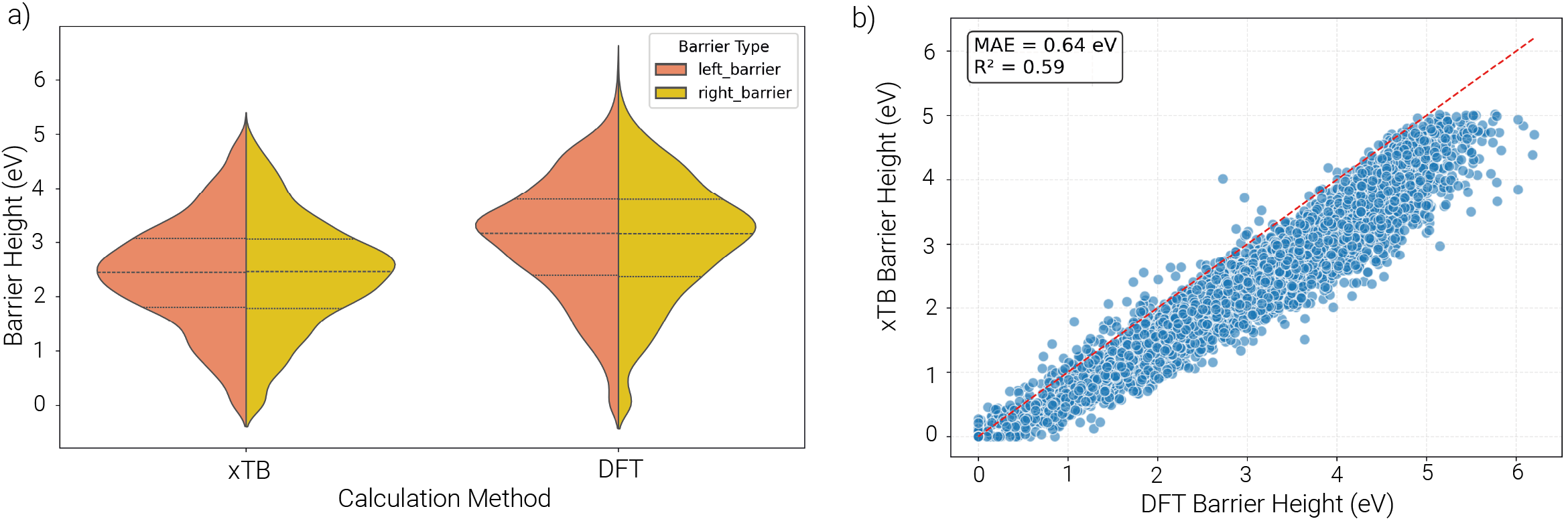}
    \caption{Comparison between barrier heights (n = 4,025) calculated using xTB vs. DFT for all configurations of the linear interpolation dataset. a) Violin plot: xTB underestimates both left and right HAT reaction barriers. b) xTB underestimates barrier heights for most systems.} 
    \label{fig:SI_barriers_violin_parity}
\end{figure}

\section{Comparative analysis of GNNs}

\begin{figure}[H]
    \centering
    \includegraphics[width=15 cm]{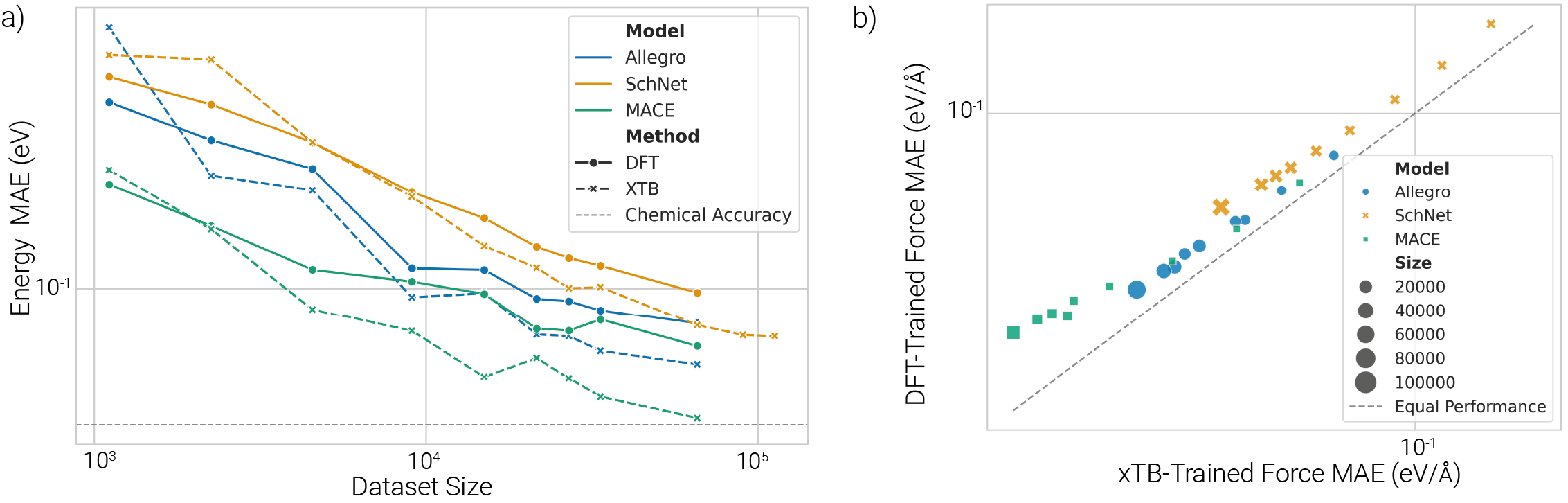}
    \caption{Scaling behavior of GNNs: a)  Test set energy MAE vs. training dataset size. b) DFT-trained force MAE is higher than xTB-trained force MAE for all trained models. } 
    \label{fig:SI_scaling_law}
\end{figure}

\begin{figure}[H]
    \centering
    \includegraphics[width=10 cm]{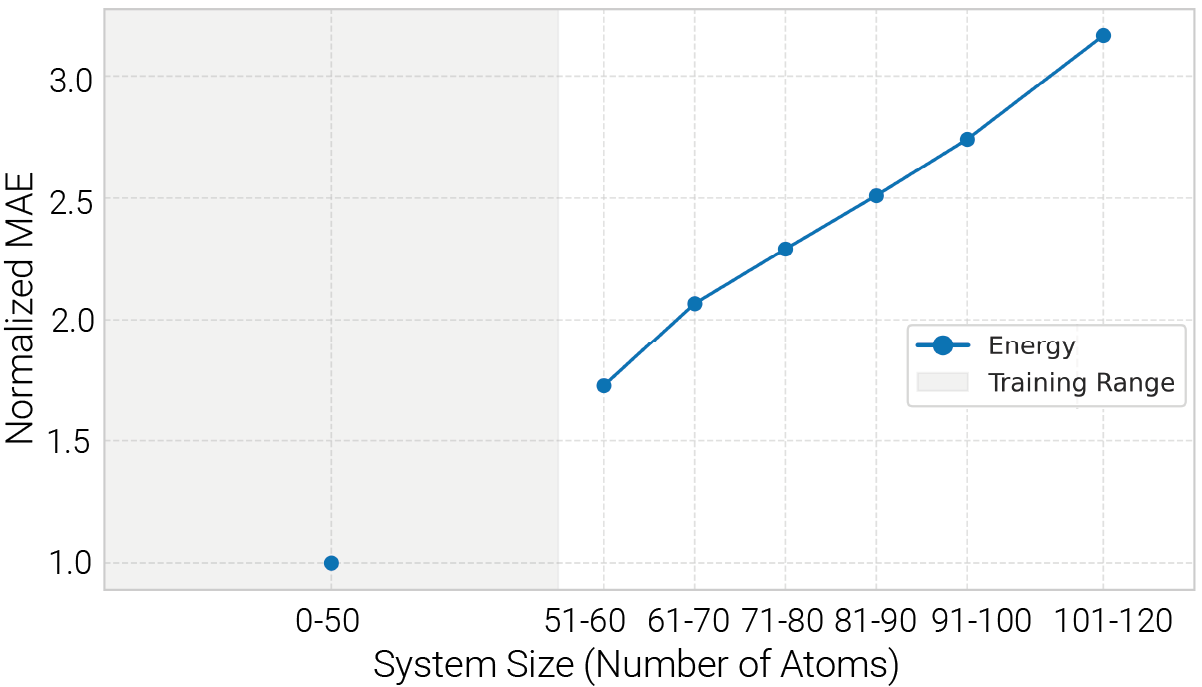}
    \caption{Transferability of MACE to different system sizes: Energy MAEs and per-atom energy MAEs vs. atom count. Energy MAE increases with system size, hinting at additive errors. } 
    \label{fig:SI_barriers_transf}
\end{figure}

\section{Final model performance}
\begin{table}[H]
\centering
\caption{Test error of models trained on 65,514 xTB configurations and tested on 6,836 unseen configurations and 2,164 barrier evaluations.}
\label{tab:model_comparison_xtb}
\begin{tabular}{lccc}
\toprule
\textbf{Model} & \textbf{Energy MAE (meV)} & \textbf{Force MAE (meV/\AA)} & \textbf{Barrier MAE (meV)} \\
\midrule
SchNet & 78 & 43 & 78 \\
Allegro & 60 & 30 & 58  \\
MACE & \textbf{42} & \textbf{18} & \textbf{39}  \\
\bottomrule
\end{tabular}
\end{table}

\end{document}